\documentclass[12pt]{article}
\usepackage[dvips]{graphicx}
\usepackage{amsfonts,amsmath,epsfig,amssymb,array,enumerate} 
\usepackage[subnum]{cases}
\usepackage[latin1]{inputenc}
\usepackage[french]{babel}
\usepackage[sort, numbers]{natbib}
\newcommand{\bt}{\mathbf{t}}

\newcommand{\by}{\mathbf{y}}

\newcommand{\bz}{\mathbf{z}}

\newcommand{\bsw}{\boldsymbol{w}}

\newcommand{\cO}{\mathcal{O}}

\newcommand{\bsbeta}{\boldsymbol{\beta}}

\newcommand{\bstheta}{\boldsymbol{\theta}}

\newcommand{\bsPsi}{\boldsymbol{\Psi}}

\newcommand{\R}{\mathbb{R}}
	
\newcommand{\N}{\mathcal{N}}
\newcommand{\M}{\mathcal{M}}

\begin{document}
\renewcommand{\refname}{Bibliographie}
\begin{center}
{\Large
	{\sc
		 Classification automatique de données temporelles en classes ordonnées}
}
\bigskip

Faicel Chamroukhi$^{1}$,   Allou Samé$^1$, Gérard Govaert$^2$, Patrice Aknin$^1$

\medskip
{\it

$^1$UPE, IFSTTAR, GRETTIA\\
2 rue de la Butte Verte, 93166 Noisy-le-Grand Cedex\\
$^2$Université de Technologie de Compiègne\\
Laboratoire HEUDIASYC, UMR CNRS 6599 \\
BP 20529, 60205 Compiègne Cedex\\
}
\end{center}
\bigskip
\noindent \textbf{Résumé} : Cet article propose une méthode de classification automatique de données temporelles en classes ordonnées. Elle se base sur les modèles de mélange et sur un processus latent discret,  qui permet d'activer successivement les classes. La classification peut s'effectuer en maximisant la vraisemblance via l'algorithme EM ou en optimisant simultanément les paramètres et la partition par l'algorithme CEM. Ces deux algorithmes peuvent être vus comme des alternatives à l'algorithme de Fisher, qui permettent d'améliorer son temps de calcul.

\noindent \textbf{Mots clés} : classification, régression, processus latent, algorithme EM, algorithme CEM.
\medskip

\noindent\textbf{Abstract} :  This paper proposes a new approach for classifying temporal data into ordered clusters. It is based on a specific model  governed by a discrete latent process which successively activates  clusters. The classification can be performed by maximizing the likelihood via the EM algorithm or by maximizing the complete-data likelihood via the  CEM algorithm. These two algorithms can be seen as alternatives to Fisher algorithm that reduce its running time.

\section{Introduction} 
L'analyse de séquences ou de séries temporelles est un sujet d'importance en  traitement du signal et dans de nombreux domaines applicatifs. Très souvent, le problème étudié se ramène à celui de la classification automatique sous une contrainte d'ordre sur les classes (classes ordonnées dans le temps). Dans ce contexte, l'objectif est de trouver, à partir d'un échantillon de données temporelles, une partition en $K$ classes ordonnées en optimisant un critère numérique.  
L'algorithme de Fisher \cite{fisher} fournit une solution optimale via un algorithme de programmation dynamique qui peut s'avérer coûteux en temps de calcul, surtout pour des échantillons de grande taille. 
Nous proposons dans cet article une approche différente pour résoudre ce problème. Celle-ci s'appuie sur un processus latent permettant d'activer successivement les classes, et sur les mélanges de lois. Deux méthodes peuvent être utilisées pour estimer la partition. La première se base sur l'algorithme EM \cite{dlr}  pour l'optimisation de la vraisemblance et la deuxième repose sur l'algorithme CEM \cite{celeuxetgovaert92_CEM} permettant d'optimiser la  vraisemblance classifiante. Nous montrons que notre approche constitue une alternative pertinente au partitionnement de Fisher, qui requiert  des temps de calculs moins élevés.
 
Dans la section 2 nous décrivons la méthode de partitionnement par l'algorithme de Fisher. Ensuite, nous introduisons dans la section 3 l'approche proposée et les deux algorithmes de classification associés. La section 4 montre les performances de la méthode proposée sur des données simulées.

\section{Segmentation optimale par l'algorithme de Fisher}
 
On suppose disposer d'un échantillon de données temporelles $((y_1,t_1),\ldots,(y_n,t_n))$, où $y_i$ désigne une variable aléatoire scalaire dépendante observée à l'instant $t_i$. Une partition en $K$ segments ordonnés suivant le temps sera notée $P_{n,K} = \left(\{y_i|i\in I_1\},\ldots,\{y_i|i\in I_K\}\right)$, où $I_k = (\xi_{k},\xi_{k+1}]$ représente les indices des observations de la classe $k$ $(k=1,\ldots,K)$ avec $\xi_1=0$ et $\xi_{K+1}=n$. 

\subsection{Cas général} 
L'algorithme de Fisher \cite{fisher} est un algorithme de programmation dynamique permettant de calculer la partition optimale $P_{n,K}$ en minimisant un critère additif suivant les classes, qui peut donc s'écrire $C (P_{n,K}) = \sum_{k=1}^K D(I_k)$, où $D (I_k)$ est le co\^ut  (appelé aussi diamètre) de la classe définie sur l'intervalle de temps $I_k$.  L'additivité de ce critère permet de l'optimiser d'une manière exacte en s'appuyant sur la programmation dynamique. Notons que dans la version standard de l'algorithme de Fisher, le diamètre $D (I_k)$ utilisé est l'inertie de la classe associée à $I_k$, qui s'écrit $D(I_k) = \sum_{i\in I_k} (y_i - a_k)^2$ avec $a_k = \sum_{i\in I_k}y_i/{\#I_k}$. Dans ce cas particulier,  chaque classe est représentée par sa moyenne qui est un scalaire dans le cas d'observations appartenant à $\R$. Les variances sont supposées égales pour toutes les classes.

\subsection{Extension au cas de classes polynomiales}

Dans les applications en traitement du signal notamment, il est souvent utile  de représenter une classe, non pas par un scalaire (fonction constante du temps), mais par un polynôme. L'algorithme de Fisher peut être aussi utilisé dans ce cas en considérant le diamètre $D(I_k) = \sum_{i\in I_k} (y_i - \bsbeta^T_k \bt_i)^2$ où le vecteur $\bsbeta_k=(\beta_{k0},\ldots,\beta_{kp})^T$ de $\R^{p+1}$ désigne l'ensemble des coefficients d'un polynôme de degré $p$ associé à la $k$ième classe et $\bt_i=(1,t_i,t_i^2,\ldots,t_i^p)^T$ son vecteur de monômes
associés. Plus généralement, l'algorithme de Fisher consiste à maximiser dans ce cas la somme des log-vraisemblances  des $K$ sous-modèles de régression relatifs aux $K$ classes, qui est un critère  additif sur les $K$ classes.  
 La procédure de programmation dynamique dans ce cas est réalisée avec une complexité temporelle en $\cO(K p^2 n^2)$. Cette complexité, bien qu'elle soit efficace pour des échantillons de petites tailles, peut s'avérer très coûteuse pour les échantillons de grande tailles. Dans la section suivante, nous proposons un alternative plus rapide qui s'inspire des modèles de mélange  de lois.

\section{Approche générative de segmentation en classes ordonnées et algorithme EM}
\subsection{Modèle}
\begin{figure}[!h]
\begin{minipage}{.43\linewidth}
\caption{Évolution de $\pi_{1}(t_i;\bsw)$ au cours du temps pour différentes valeurs de $\lambda_1$ avec $\gamma_1 = -2$ et $K=2$.} 
\end{minipage} 
\begin{minipage}{.3\linewidth}
  \centering
  \includegraphics[scale=.2]{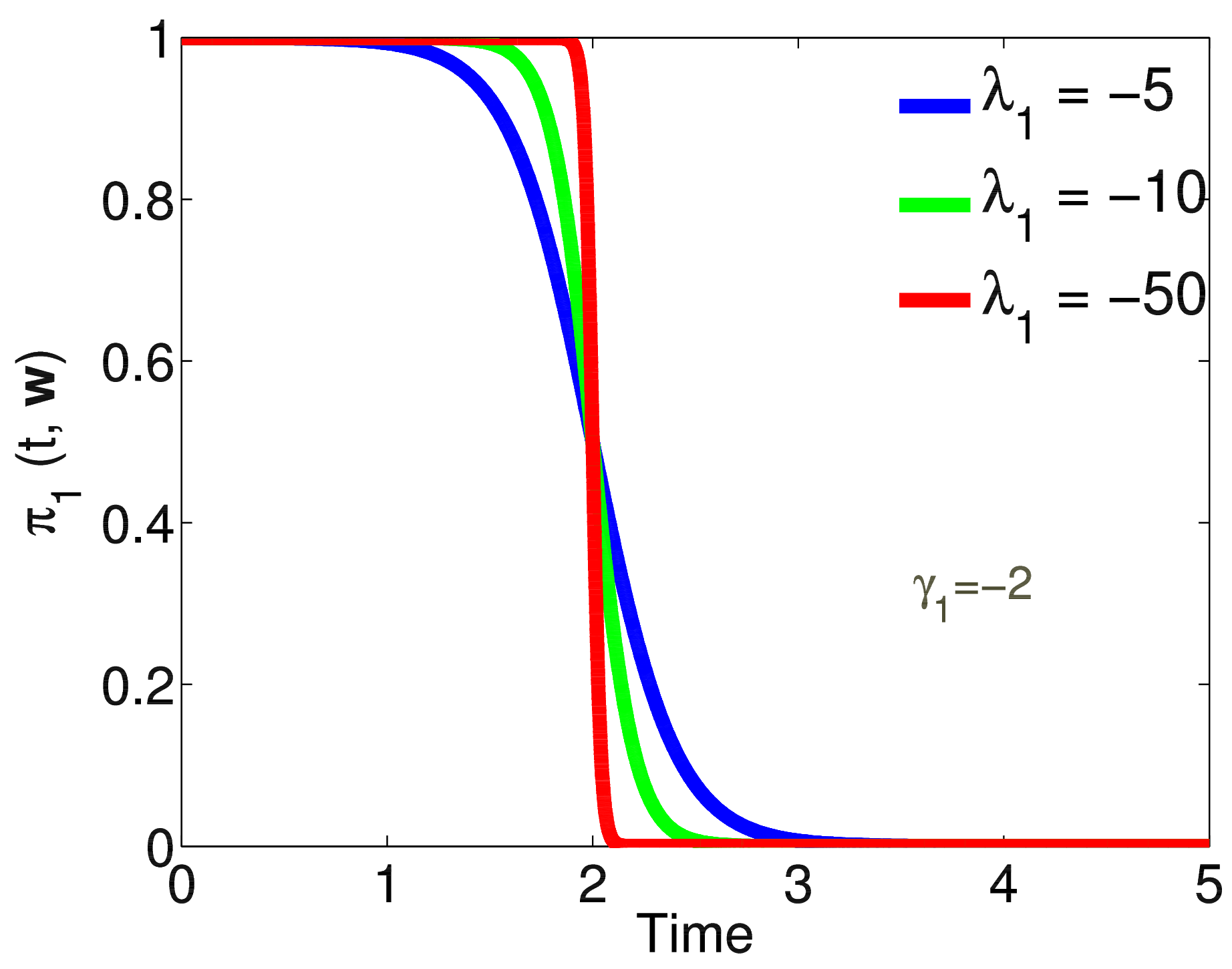}
 \end{minipage} \hfill
 \label{fig: illustration de la fonction logistique}
\end{figure}

L'approche de classification proposée repose sur un processus latent qui permet de spécifier, pour chaque point $y_i$, une classe $z_i \in \{1,\ldots,K\}$. Chaque classe $k$ est représentée de manière plus générale par un polynôme. Le modèle se définit comme suit:
\begin{equation}
y_i = \sum_{k=1}^K z_{ik} \bsbeta_k^T\bt_i  + \sigma_{k} \epsilon_i, \quad (i=1,\ldots,n)
\label{eq:  rhlp regression model}
\end{equation}
où $z_{ik}$ est la variable qui vaut 1 si $z_i=k$ et  0 sinon, les $\epsilon_i$ sont des bruits gaussiens $i.i.d$ centrés réduits et $\sigma_{k}$ un réel positif. La variable $z_i$ contrôle le passage d'une classe à l'autre au cours du temps. On suppose que la séquence de ces variables latentes $\bz = (z_1, \ldots, z_m)$ est un processus logistique latent selon lequel les variables $z_i$, conditionnellement aux instants $t_i$, sont supposées être générées indépendamment suivant la loi multinomiale $\M(1,\pi_1(t_i;\bsw),\ldots,\pi_K(t_i;\bsw))$, où la probabilité $\pi_k(t_i;\bsw)$ de la classe $k$ est définie par la fonction logistique  
$$\pi_k(t_i;\bsw)=p(z_i=k|t_i;\bsw)=\frac{\exp(\lambda_k(t_i+\gamma_k))}{\sum_{\ell=1}^K \exp(\lambda_\ell(t_i +\gamma_\ell))}$$
 paramétrée par $\bsw=(\lambda_1,\gamma_1,\ldots,\lambda_K,\gamma_K)$ de $\R^{2K}$. 
L'utilisation des fonctions logistiques comme probabilités des classes nous permet notamment d'estimer une partition ordonnée. En effet,  d'après leur définition, on peut facilement vérifier que $\log \frac{\pi_{k}(t_i;\bsw)}{\pi_{\ell}(t_i;\bsw)}$ est linéaire en $t$; par conséquent, la partition obtenue en maximisant les probabilités $\pi_k(t_i;\bsw)$ est convexe et les segments obtenus sont ordonnés suivant le temps. En outre, comme  l'illustre la Figure \ref{fig: illustration de la fonction logistique}, l'approche de partitionnement par l'algorithme de Fisher  est un cas limite de l'approche proposée. Pour chaque probabilités $\pi_k(t_i;\bsw)$, il suffit de tendre $\lambda_k$ vers $-\infty$. 

On peut montrer que conditionnellement à $t_i$, la variable aléatoire $y_i$ est distribuée suivant le mélange de densités normales $p(y_i|t_i;\bsPsi)=\sum_{k=1}^K \pi_k(t_i;\bsw) \N \big(y_i;\bsbeta^T_k\bt_i,\sigma_k^2 \big)$  où $\N(\cdot;\mu,\sigma^2)$ désigne la fonction de densité d'une loi normale d'espérance $\mu$ et de variance $\sigma^2$ et $\bsPsi=(\bstheta, \bsw)$ représente l'ensemble des paramètres du modèle.  Les deux paragraphes suivants présentent deux méthodes d'estimation des paramètres et de la partition.

\subsection{1$^{\text{ère}}$ approche : estimation des paramètres par EM et calcul de la partition ordonnée}

Comme la densité de la loi conditionnelle de $y_i$ s'écrit sous la forme d'un mélange, nous exploitons le cadre de l'algorithme EM \cite{dlr} pour estimer ses paramètres. La log-vraisemblance à maximiser s'écrit $\log p(\by|\bt;\bsPsi)= \sum_{i=1}^{n}\log\sum_{k=1}^K\pi_{k}(t_i;\bsw)\mathcal{N}\big(y_i;\bsbeta^T_k\bt_i,\sigma_k^2\big)$. Cette maximisation ne pouvant pas être effectuée analytiquement, nous nous appuyons sur l'algorithme EM \cite{dlr} pour l'effectuer. L'algorithme EM, dans cette situation, itère à partir d'un paramètre initial  $\bsPsi^{(0)}$ les deux étapes suivantes jusqu'à la convergence :

\medskip
\noindent \textbf{Étape E} : Cette étape consiste à calculer l'espérance de la log-vraisemblance complétée \linebreak  
{\small $\log p(\by,\bz|\bt;\bsPsi) =\sum_{i=1}^{n}\sum_{k=1}^K  z_{ik} \log
[\pi_{k}(t_i;\bsw)\mathcal{N} (y_i;\bsbeta^T_k\bt_{i},\sigma_k^2)]$},  
conditionnellement aux observations et au paramètre courant
$\bsPsi^{(q)}$. Dans notre situation, cette espérance conditionnelle s'écrit:  
{\small $Q(\bsPsi,\bsPsi^{(q)}) = \sum_{i=1}^{n}\sum_{k=1}^K\tau^{(q)}_{ik}\log [\pi_{k}(t_i;\bsw)\mathcal{N}(y_i;\bsbeta^T_k\bt_{i},\sigma_k^2)]$}
où
$ \tau^{(q)}_{ik} = p(z_{ik}=1|y_i,t_i;\bsPsi^{(q)})  =  \frac{\pi_{k}(t_i;\bsw^{(q)})\mathcal{N}(y_i;\bsbeta^{T(q)}_k\bt_{i},\sigma_k^{2(q)})}{\sum_{\ell=1}^K\pi_{\ell}(t_i;\bsw^{(q)})\mathcal{N}(y_i;\bsbeta^{T(q)}_{\ell}\bt_{i},\sigma_\ell^{2(q)})}$
est la probabilité a posteriori de la classe $k$.  

\medskip

\noindent \textbf{Étape M} : Cette étape de mise à jour consiste à calculer le paramètre $\bsPsi^{(q+1)}$ qui maximise $Q(\bsPsi,\bsPsi^{(q)})$ par rapport à $\bsPsi$. La quantité $Q(\bsPsi,\bsPsi^{(q)})$ s'écrit sous la forme d'une somme de  deux quantités : $Q_1(\bsw)=\sum_{i=1}^{n}\sum_{k=1}^K \tau^{(q)}_{ik}\log \pi_{k}(t_i;\bsw)$ et $Q_2(\{\bsbeta_k,\sigma_k^2\}) = \sum_{i=1}^{n}\sum_{k=1}^{K} \tau^{(q)}_{ik}\log \N (y_i;\bsbeta^T_k\bt_{i},\sigma_k^2)$ pouvant être  maximisées séparément. La maximisation de $Q_2(\{\bsbeta_k,\sigma_k^2\})$ se ramène à résoudre analytiquement $K$ problèmes de moindres-carrés ordinaires pondérés par les probabilités à posteriori $\tau^{(q)}_{ik}$. La maximisation de $Q_1$ par rapport à $\bsw$ est un problème convexe de régression logistique multinomial pondéré par les $\tau^{(q)}_{ik}$ qui est résolu par l'algorithme IRLS  \cite{irls}.  A la convergence de l'algorithme EM, une partition en classes ordonnées est ensuite calculée à partir des probabilités logistiques.  

\medskip
L'algorithme EM proposé opère avec une complexité temporelle de $\cO(NM K^3p^2 n)$, où $N$ est le nombre d'itérations de l'algorithme EM et $M$ est le nombre moyen d'itérations requis par l'algorithme IRLS.  Ainsi, le ratio entre la complexité de l'algorithme proposé et celle de l'algorithme de Fisher est de $\frac{N M K^2}{n}$. En pratique $K$ étant souvent petit par rapport à $n$, l'algorithme EM permet d'améliorer le temps de calcul. En outre, on peut adopter une version classifiante encore plus rapide en se basant sur l'algorithme CEM décrit ci-après.

\vspace*{-.5cm}
\subsection{2$^{\text{ème}}$ approche : estimation directe de la partition et des paramètres (algorithme CEM)}

Nous proposons également une version classifiante qui s'appuie sur l'algorithme CEM \cite{celeuxetgovaert92_CEM}, dans laquelle on estime simultanément, à chaque itération, les paramètres $\bsPsi$ du modèle  et la partition définie par les classes $\bz$ en maximisant cette fois ci la log-vraisemblance complétée (classifiante) $\log p(\by,\bz|\bt;\bsPsi)$ simultanément par rapport à $\bsPsi$ et $\bz$. Nous la noterons par la suite $C(\bsPsi,\bz)$. 
Dans notre cas, cela consiste à  intégrer une étape de classification (étape C) entre les étapes E et M,  dans laquelle on estime à chaque itération une partition ordonnée en maximisant les probabilités logistiques.  Ainsi, étant donné un vecteur paramètre initial $\bsPsi^{(0)}$, une itération de l'algorithme  CEM  proposé peut être résumée comme suit:

\medskip
\noindent \textbf{Etape E} : Calculer les probabilités a posteriori $\tau^{(q)}_{ik}$  

\medskip
\noindent \textbf{Etape C} :  Estimer les classes $\bz^{(q)}$ en affectant chaque observation $y_i$ au segment maximisant les probabilités logistiques:  
$z^{(q)}_{i} = \arg \max_k  \tau_{ik}^{(q)}$

\medskip
\noindent \textbf{Etape M} : Mettre à jour les paramètres : $\bsPsi^{(q+1)} = \arg \max_{\bsPsi}   C(\bsPsi,\bz^{(q)}).$ 

\medskip
A la convergence, on obtient une partition en classes ordonnées et un vecteur paramètre. 
 Notons que les expressions de mise à jour des paramètres à l'étape M sont similaires à celles de l'algorithme EM et consistent simplement à remplacer les $\tau_{ik}$ par les $z_{ik}$. 
 En pratique, l'algorithme CEM converge plus rapidement que l'algorithme EM mais les estimations fournies peuvent être biaisées si les classes sont mal séparées puisque  la mise à jour des paramètres s'effectue à partir d'un échantillon tronqué \cite{celeuxetgovaert92_CEM}.

\section{Expérimentations sur des données simulées}

\begin{figure}[!h]
\begin{minipage}{.45\linewidth}
\caption{Exemple de données simulées pour la situation 1 (gauche) et la situation 2 (droite) avec $n=300$, $\sigma_1 =1$, $\sigma_2 =1.5$ et $\sigma_3 =2$}
\end{minipage}
\begin{minipage}{.55\linewidth}
  \centering
  \includegraphics[width=3.8cm,height=2.8 cm]{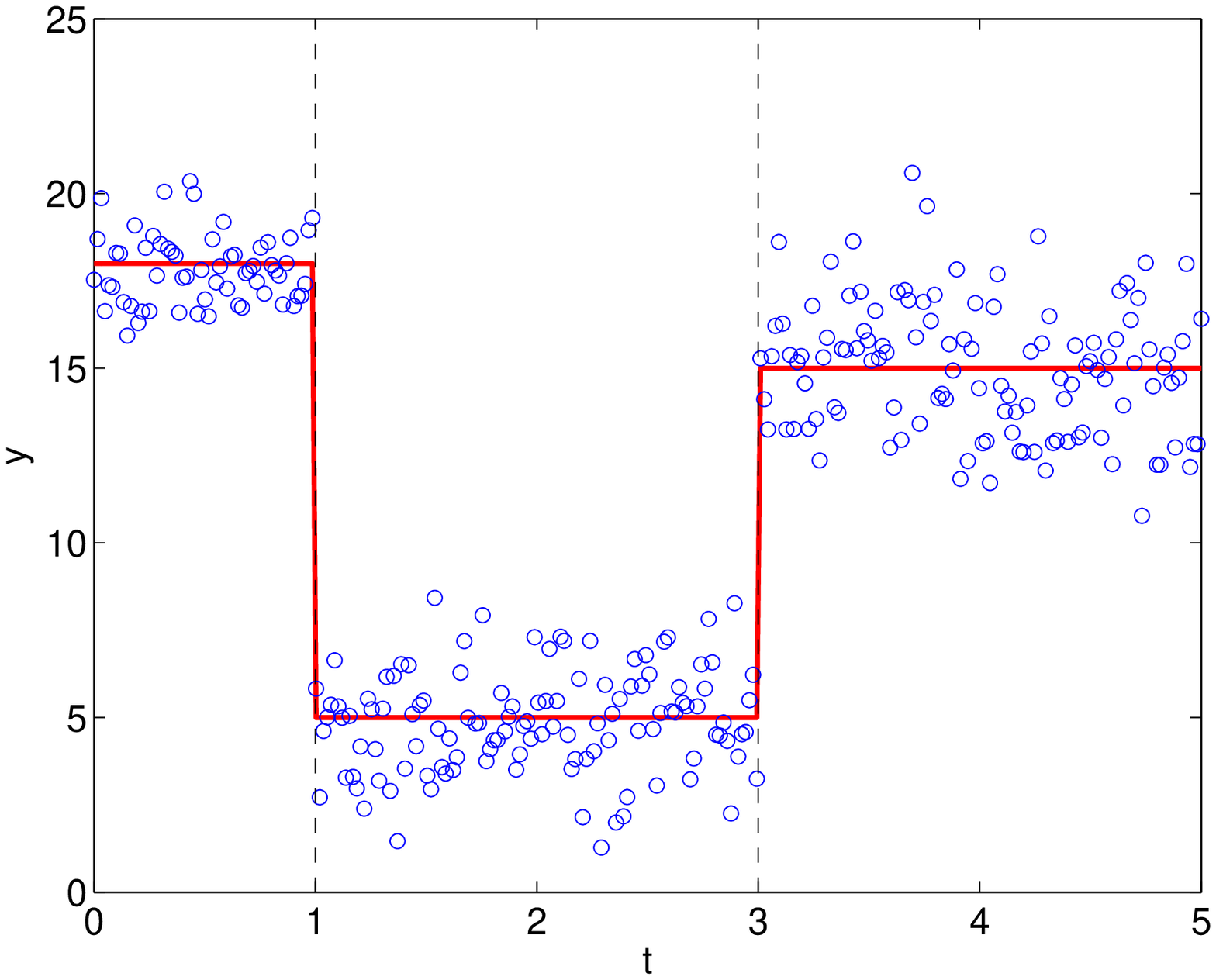}
    \includegraphics[width=3.8cm,height=2.8 cm]{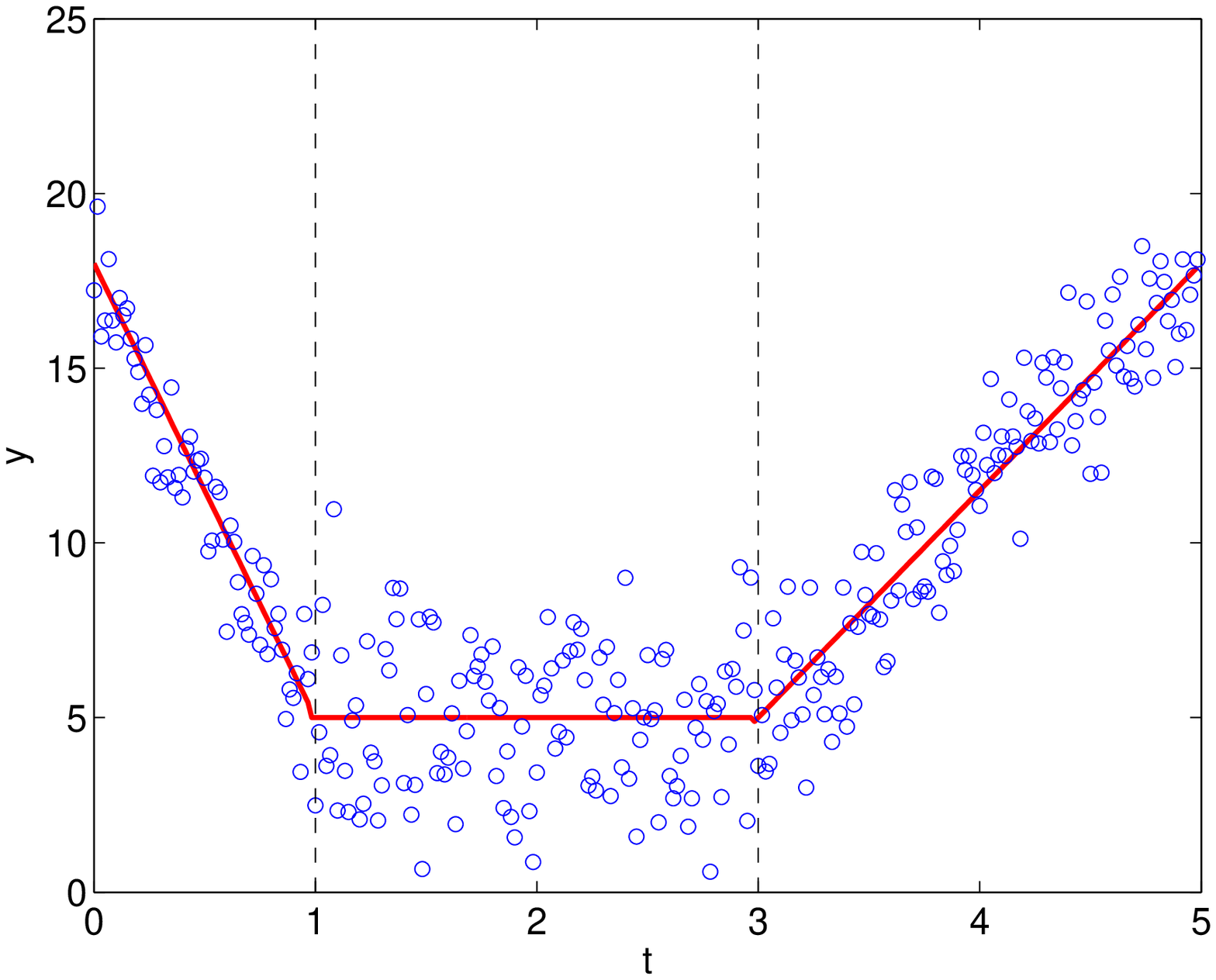}  
\end{minipage} \hfill
\label{fig: exemple de simus}
\end{figure}

Dans cette partie nous évaluons l'approche proposée en utilisant des données temporelles simulées. Pour ce faire, nous comparons les algorithmes EM, CEM et l'algorithme de Fisher. La qualité des estimations est quantifiée à travers l'erreur de segmentation et le temps de calcul. Chaque jeu de données est généré en ajoutant un bruit gaussien centré à trois segments ordonnées au cours du temps et régulièrement échantillonnés sur l'intervalle temporel $[0;5]$. Les instants de transitions sont $t_1=1$ et $t_2 = 3$.  La Figure \ref{fig: exemple de simus} montre un exemple de données simulées pour chacune des deux situations considérées.

Les algorithmes ont été appliqués avec $(K=3,p=0)$  pour la situation 1 et $(K=3,p=1)$ pour la situation 2. Les erreurs de segmentation obtenues avec les trois algorithmes pour les deux situations sont quasi-identiques et sont données dans le tableau \ref{table. erreurs de classif}.  
\vspace*{-.5cm}
\begin{table}[!h]
\centering
{\scriptsize \begin{tabular}{lcccccccccc}
\hline
 &  Algorithme de Fisher & Algorithme EM & Algorithme CEM \\
\hline 
Situation 1 &     0  &   0  & 0 \\
Situation 2 &     3.02  &   2.85   & 3.00  \\ 
\hline
\end{tabular}}
\caption{ {\small Taux d'erreur de classification moyen en \% pour $n=500$.}}
\label{table. erreurs de classif}
\end{table}

Le tableau \ref{table. temps de calcul} montre que l'algorithme EM et l'algorithme CEM pour l'approche proposée, ont des temps de calculs très réduits et qui augmentent très légèrement avec la taille d'échantillon. Cependant, le temps de calcul de l'algorithme de Fisher augmente considérablement avec la taille d'échantillon.
\begin{table}[!h]
\centering
{\scriptsize \begin{tabular}{l|ccc| ccccccc}
\hline
 &  &   Situation 1 &  & & situation 2 \\
$n$    &  Algo. de Fisher & Algo. EM & Algo. CEM    & Algo. de Fisher & Algo. EM & Algo. CEM\\
\hline 
100 &     0.1894  &   0.0800  & 0.0730 		&     0.1908  &  0.1720  &  0.0520\\
300 &     1.8153  &   0.2100  & 0.1700   	&     2.6498  &  0.2920  &  0.1720\\
500 &     5.3657  &   0.2350  & 0.2130 		&     7.6266  &  0.4360  &  0.2540\\
700 &    11.2800  &   0.2730  & 0.2640  	&    16.7204  &  0.5260  &  0.3980\\
1000 &    24.7356 &   0.2950  &  0.2840	 	&    44.7683  &  0.6080  &  0.4140\\
1500 &    61.0176 &   0.4100  &  0.3770 	&   107.0891  &  1.1320  &  0.5780\\
2000 &   118.5632 &   0.5500  &  0.4630 	&   226.4174  &  1.1340  &  0.7900\\
3000 &   310.4002 &   0.8600  &  0.6740  	&   633.5318  &  1.1920 &  1.0100\\
\hline
\end{tabular}}
\caption{ {\small Temps de calculs moyens en secondes, en fonction de la taille d'échantillon $n$.}}
\label{table. temps de calcul}
\end{table}

 \vspace*{-.8 cm}

\section{Conclusion}

Dans cet article nous avons proposé une approche de classification automatique de données temporelles en classes ordonnées. Cette approche peut être vue comme une version générative du modèle de régression par morceaux qui s'appuie sur l'algorithme de Fisher. L'algorithme EM et l'algorithme CEM proposés pour la classification se distinguent surtout par leur rapidité comparée à l'algorithme de Fisher.

\vspace*{-.3cm}
{

}
\end{document}